\documentclass{article}
\usepackage{ijcai20}

\usepackage{times}

\usepackage{soul}
\usepackage{url}
\usepackage[hidelinks]{hyperref}
\usepackage[utf8]{inputenc}
\usepackage[small]{caption}

\usepackage{graphicx}
\usepackage{subfig}

\usepackage{amsmath}
\usepackage{amssymb}
\usepackage{multirow}
\usepackage{booktabs}
\usepackage{dblfloatfix}
\usepackage{epstopdf}
\title{What and Where to Translate: Local Mask-based Image-to-Image Translation}

\author{
Wonwoong Cho$^1$\footnote{Contact Author}\and
Seunghwan Choi$^2$\and
Junwoo Park$^2$\and
David Keetae Park$^3$\and
\\
Tao Qin$^4$\and
Jaegul Choo$^1$\\
\affiliations
$^1$Korea University\\
$^2$Korea Advanced Institute of Science and Technology\\
$^3$Columbia University\\
$^4$Microsoft Research Asia
}

\begin{document}

\maketitle

\begin{abstract}
Recently, image-to-image translation has obtained significant attention. Among many, those approaches based on an exemplar image that contains the target style information has been actively studied, due to its capability to handle multimodality as well as its applicability in practical use. However, two intrinsic problems exist in the existing methods: what and where to transfer. First, those methods extract style from an entire exemplar that includes noisy information, which impedes a translation model from properly extracting the intended style of the exemplar. That is, we need to carefully determine what to transfer from the exemplar. Second, the extracted style is applied to the entire input image, which causes unnecessary distortion in irrelevant image regions. In response, we need to decide where to transfer the extracted style. In this paper, we propose a novel approach that extracts out a local mask from the exemplar that determines what style to transfer, and another local mask from the input image that determines where to transfer the extracted style. The main novelty of this paper lies in (1) the highway adaptive instance normalization technique and (2) an end-to-end translation framework which achieves an outstanding performance in reflecting a style of an exemplar. We demonstrate the quantitative and qualitative evaluation results to confirm the advantages of our proposed approach.
\end{abstract}

\section{Introduction}
\label{sec:intro}
Unpaired image-to-image translation, in short, image translation, based on generative adversarial networks (GANs)~\cite{goodfellow2014generative} aims to transform an input image from one domain to another, without using paired data between different domains~\cite{Zhu_2017,liu2017unsupervised,kim2017learning,StarGAN2018,Bahng_2018_ECCV}. 
Such an unpaired setting is inherently multimodal, since a single input image can be mapped to multiple different outputs within a target domain. For example, when translating the hair color of a given image into a blonde, the detailed hair region (e.g., upper vs. lower, and partial vs. entire) and detailed color (e.g., dark vs. light blonde) may vary. 

Previous studies have achieved such multimodal outputs by adding a random noise sampled from a pre-defined prior distribution~\cite{zhu2017toward} or taking a user-selected exemplar image as additional input, which contains the detailed information of an intended target style~\cite{chang2018pairedcyclegan}. Recent studies~\cite{lin2018conditional} including MUNIT~\cite{Huang_2018_ECCV} and DRIT~\cite{Lee_2018_ECCV} utilize these two approaches, showing the state-of-the-art performance by separating (i.e., disentangling) content and style information of a given image through two different encoder networks.

However, existing exemplar-based methods have several limitations as follows. 
First, those methods do not pay attention to what target style to transfer from the exemplar. Instead, they simply extract style information from the entire region of a given exemplar, while it is likely only the style of a sub-region of the exemplar should be transferred. Thus, the style of the entire exemplar will tend to be noisy due to the irrelevant regions with respect to the target attribute to transfer. It gives rise to a degradation of the model, particularly in reflecting only the relevant style contained in an exemplar.

Suppose we translate the hair color of an image using an exemplar image. Since the hair color information is available only in the hair region of an image, the style information extracted from the entire region of the exemplar may contain the irrelevant information (e.g., the color of the wall and the texture pattern of the floor), which should not be reflected in the intended image translation.
In the end, the noisy style results in erroneous translations in mirroring the hair color of the exemplar, as illustrated in Fig.~\ref{fig:comparison with baseline methods}.

Second, previous methods do not distinguish different regions of the input image. Even though particular regions should be kept as it is during translation, those methods simply transfer the extracted style to the entire region of the input image to obtain the target image. Due to this issue, previous approaches~\cite{Huang_2018_ECCV,Lee_2018_ECCV} often distort the irrelevant regions of an input image such as the background. That is, we should be aware of where to transfer for the input image.


To tackle these issues, we propose a novel, \underline{LO}cal \underline{M}ask-based \underline{I}mage \underline{T}ranslation approach, called LOMIT, which generates a local, pixel-wise soft binary mask of an exemplar (i.e., the source region from which to extract out the style information) to identify what style to transfer and that of an input image to identify where to translate (i.e., the target region to which to apply the extracted style). 

Our algorithm shares the similar high-level idea as recent approaches~\cite{ma2019exemplar,mejjati2018unsupervised,Zhang2018SaGAN} that have leveraged an attention mask in image translation. In those approaches, the attention mask, extracted from an input image, determines the target region to apply a translation, i.e., where to translate. However, those studies commonly deal with only single attribute translation, which limits the practical use of the translation framework. Moreover, we expand those approaches by additionally exploiting a mask for an exemplar, so that LOMIT enables a user to decide where to translate as well as what to be transferred. 





Once obtaining local masks, LOMIT extends adaptive instance normalization, using highway networks~\cite{srivastava2015highway}, which computes the weighted average of the input and the translated pixel values using the above-mentioned pixel-wise local mask values as different linear combination weights per pixel location. LOMIT has an additional advantage of being able to manipulate the computed masks to selectively transfer an intended style, e.g., choosing either a hair region (to transfer the hair color) or a facial region (to transfer the facial expression).

The effectiveness of LOMIT is evaluated with other state-of-the-art methods.


\begin{figure*}[t]
  \includegraphics[width=\linewidth]{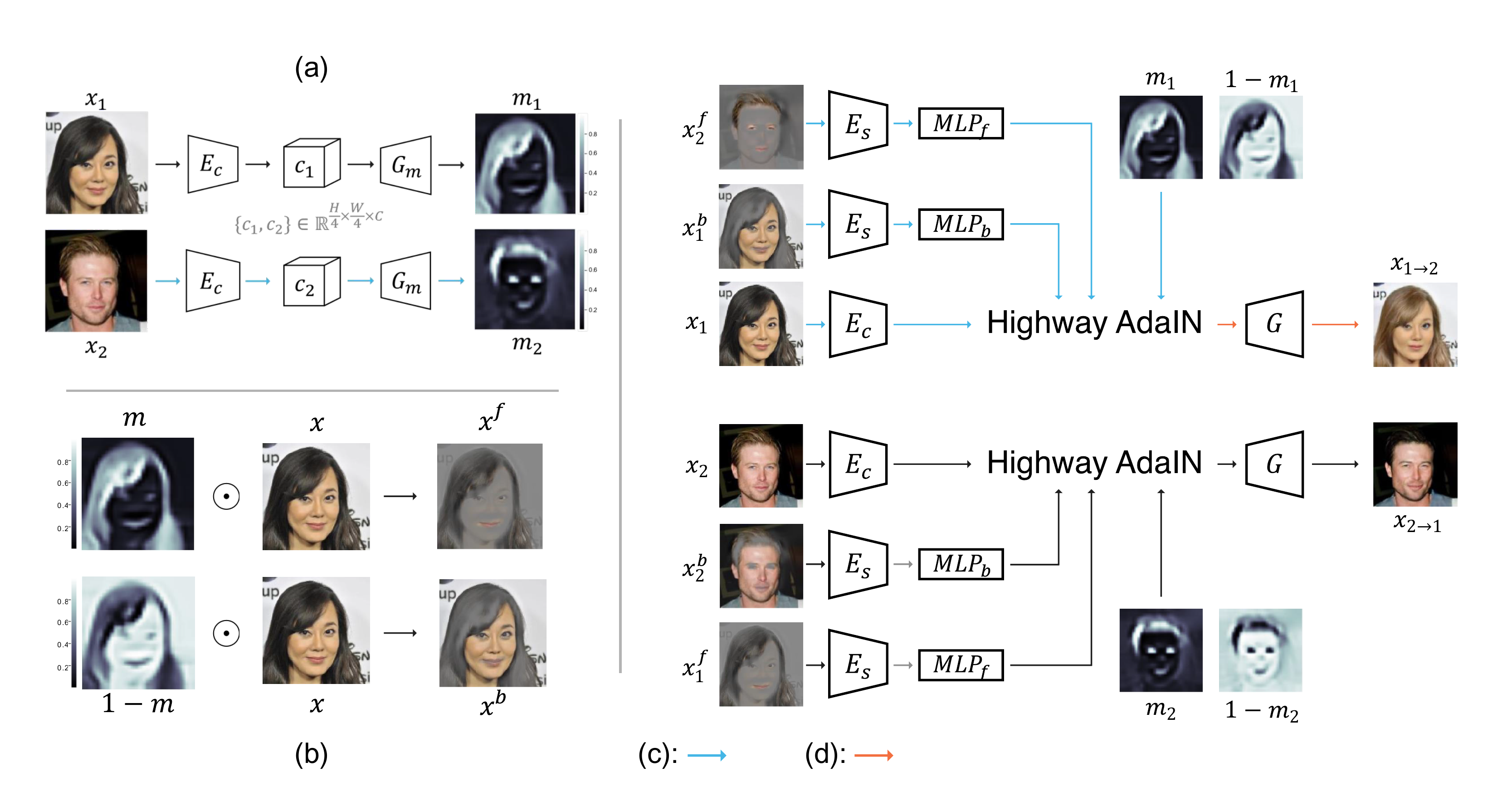}
  \vspace{-1cm}
  \caption{Image translation workflow. (a) LOMIT first generates masks for the input and the exemplar via attention networks. (b) Next, we separate each image of $x_1$ and $x_2$ into a foreground and a background regions, depending on how much each pixel is involved in image translation. (c,d) By combining the content and the background style representations from ${x_1}$ with the foreground style representation ${x_2}$, we obtain a translated image ${x_{1\rightarrow 2}}$. Note that LOMIT also learns the opposite-directional image translation ${x_{2\rightarrow 1}}$ by interchanging $x_1$ and $x_2$. Finally, LOMIT learns image translation using the cycle consistency loss from $\mathcal{X}_1\rightarrow\mathcal{X}_2\rightarrow\mathcal{X}_1$ and $\mathcal{X}_2\rightarrow\mathcal{X}_1\rightarrow\mathcal{X}_2$. 
  }
  \label{fig:overview}
\end{figure*}
\section{Basic Setting}\label{probem formulation}
We define ``content'' as common features (an underlying structure) across all domains (e.g., the pose of a face, the location and the shape of eyes, a nose, a mouth, and hair), and ``style'' as a representation of the structure (e.g., background color, facial expression, skin tone, and hair color). We assume that an image ${x}$ can be represented as ${x=c\oplus\\s}$, where ${c}$ is a content code in a content space, and ${s}$ is a style code in a style space. The operator ${\oplus}$ combines and converts the content code ${c}$ and the style code ${s}$ into a complete image ${x}$.

By considering the local mask indicating the relevant region (or simply, the foreground) to extract the style from or to apply it to, we further assume that ${s}$ is decomposed into ${s=s^f\oplus\\s^b}$, where ${s^f}$ is the style code extracted from the foreground region of the exemplar and ${s^b}$ is that from the 
The pixel-wise soft binary mask ${m}$ of an image ${x}$ is represented as a matrix with the same spatial resolution of ${x}$. Each entry of ${m}$ lies between 0 and 1, which indicates the degree of the corresponding pixel belonging to the foreground. Then, the local foreground and the background regions, $x^f$ and $x^b$ of $x$, are obtained as
\begin{equation*}\label{eq:3}
    \begin{matrix}
        x^f={m}\odot x, &
        x^b=(1-{m})\odot x,
    \end{matrix}
\end{equation*}
where ${\odot}$ indicates an element-wise multiplication.
Finally, our assumption is extended to ${x=c\oplus\\s^f\oplus\\s^b}$, where $c$, $s^f$, and $s^b$ are obtained by the content encoder $E_c$ and the style encoder $E_s$, respectively, i.e., 
\begin{equation*}
    \begin{matrix}
        \{c_x,s_x^f,s_x^b\}=\{E_c(x),E_s(x^f),E_s(x^b)\}.
    \end{matrix}
\end{equation*}
It is essential for LOMIT to properly learn to generate the local mask involved in image translation because the local mask plays a role in separating a style representation ${s}$ into ${s^f}$ and ${s^b}$. To this end, we propose to combine the mask generation networks with our novel highway adaptive instance normalization, as will be described in Section.~\ref{sec:hadain}.

\section{Local Image Translation Model}

We first denote ${x_1}\in\mathcal{X}_1$ and ${x_2}\in\mathcal{X}_2$ as images from domains $\mathcal{X}_1$ and $\mathcal{X}_2$, respectively. As shown in Fig.~\ref{fig:overview}, LOMIT converts a given image $x_1$ to $\mathcal{X}_2$ and vice versa, i.e., ${x_{1\rightarrow 2}=}$ ${G(h(E_c(x_1),E_s(x_2^f),E_s(x_1^b)))}$, and ${x_{2\rightarrow 1}=}$ ${G(h(E_c(x_2),E_s(x_1^f),E_s(x_2^b)))}$, where ${G}$ is decoder networks and ${h}$ is the local mask-based highway adaptive instance normalization layer (or in short, HAdaIN). This will be described in detail in Section.~\ref{sec:hadain}.


For a brevity purpose, we omit the domain index notation in, say, ${m={\{m_1,m_2\}}}$ and ${x=\{x_1,x_2\}}$, unless needed for clarification. 


\subsection{Local Mask Extraction}
We extract the local masks of the input and the exemplar images, as those are effectively involved in image translation. In concrete, LOMIT utilizes the local mask ${m}$ when (1) acquiring disentangled style features ${E_s(x^f), E_s(x^b)}$, and (2) specifying where to apply the style. For example, if LOMIT is conducting a hair color translation given the input image and the exemplar, our local masks should be obtained as the hair regions from two images. This is because the style to replace and transfer exist in the hair regions of the images.


As shown in Fig.~\ref{fig:overview}(a), given an image ${x}$, attention networks ${G_m}$ encode the content feature ${c}$ via the content encoder ${E_c}$. The obtained ${c}$ is then forwarded into the rest of the attention networks ${G_m}$, i.e., ${m=G_m(E_c(x))}$, where ${m}$ is the mask specifying the relevant region with respect to a target style to translate. In practice, the process applies to images in each domain independently in a similar manner, resulting in ${m_1, m_2}$.



\subsection{Highway Adaptive Instance Normalization} 
\label{sec:hadain} 
Adaptive instance normalization is an effective style transfer technique~\cite{huang2017arbitrary}. Generally, it matches the channel-wise statistics, e.g., the mean and the variance, of the activation map of an input image with those of a style image. In the context of image translation, MUNIT~\cite{Huang_2018_ECCV} extends AdaIN in a way that the target mean and the variance are obtained as the outputs of the trainable functions $\beta$ and $\gamma$ of a given style code, i.e., 
\begin{equation*}\label{eq:adain}
    \text{AdaIN}_{\mathcal{X}_{1\rightarrow 2}}(c_1,s_2)=\gamma(s_2)\left(\frac{c_1-\mu(c_1)}{\sigma(c_1)}\right)+\beta(s_2),
\end{equation*}

where each of ${\beta}$ and ${\gamma}$ is defined as a multi-layer perceptron (MLP), i.e., ${[\beta(s^f);\gamma(s^f)]=\text{MLP}_f(s^f)}$ and 
${[\beta(s^b);\gamma (s^b)]}$ ${=\text{MLP}_b(s^b)}$. Different MLPs for foreground and background are used because style information significantly differs, e.g., facial attributes vs. background texture.

As we pointed out earlier, previous approaches applied such a transformation globally across the entire region of an image, which may unnecessarily distort irrelevant regions. Hence, we formulate our local mask-based highway AdaIN (HAdaIN) as
\begin{align*}\label{eq:hadain}
    &\text{HAdaIN}_{\mathcal{X}_{1\rightarrow 2}}(m_1,c_1,s_2^f,s_1^b)= \\[0.3em] 
    &m_1~{\odot}~\text{AdaIN}_{\mathcal{X}_{1\rightarrow 2}}(c_1,s_2^f)+(1-m_1)~{\odot}~\text{AdaIN}_{\mathcal{X}_{1\rightarrow 1}}(c_1,s_1^b),
\end{align*}
where the first term corresponds to the local region of an input image translated by the foreground style, while the second corresponds to the complementary region where the original style of the input should be kept as it is.

\subsection{Training Objectives}
\paragraph{Style and content reconstruction loss.} 
The foreground style of the translated output should be close to that of the exemplar, while the background style of the translated output should be close to that of the original input image. We formulate this criteria as the following style reconstruction loss terms:
\begin{align*}
    \mathcal{L}_{s_f}^{1\rightarrow 2}&={}\mathbb{E}~[\lVert E_s(x_{1\rightarrow 2}^f)-E_s(x_2^f) \rVert_1], \\[0.3em]
    \mathcal{L}_{s_b}^{1\rightarrow 2}&=\mathbb{E}~
    [\lVert E_s(x_{1\rightarrow 2}^b)-E_s(x_1^b) \rVert_1].
\end{align*}
From the perspective of content information, the content feature of an input image should be consistent with its translated output, represented as the content reconstruction loss as
\begin{equation*}\label{recon_c}
    \mathcal{L}_{c}^{1\rightarrow 2}=\mathop{{}\mathbb{E}}[\lVert E_c(x_{1\rightarrow 2})-E_c(x_1) \rVert_1].
\end{equation*}
By encouraging the content features ${c_1}$ and ${c_{1\rightarrow 2}}$ to be the same, the content reconstruction loss maintains the content information of the input image, even after performing a translation.

\begin{figure*}
  \includegraphics[width=\linewidth]{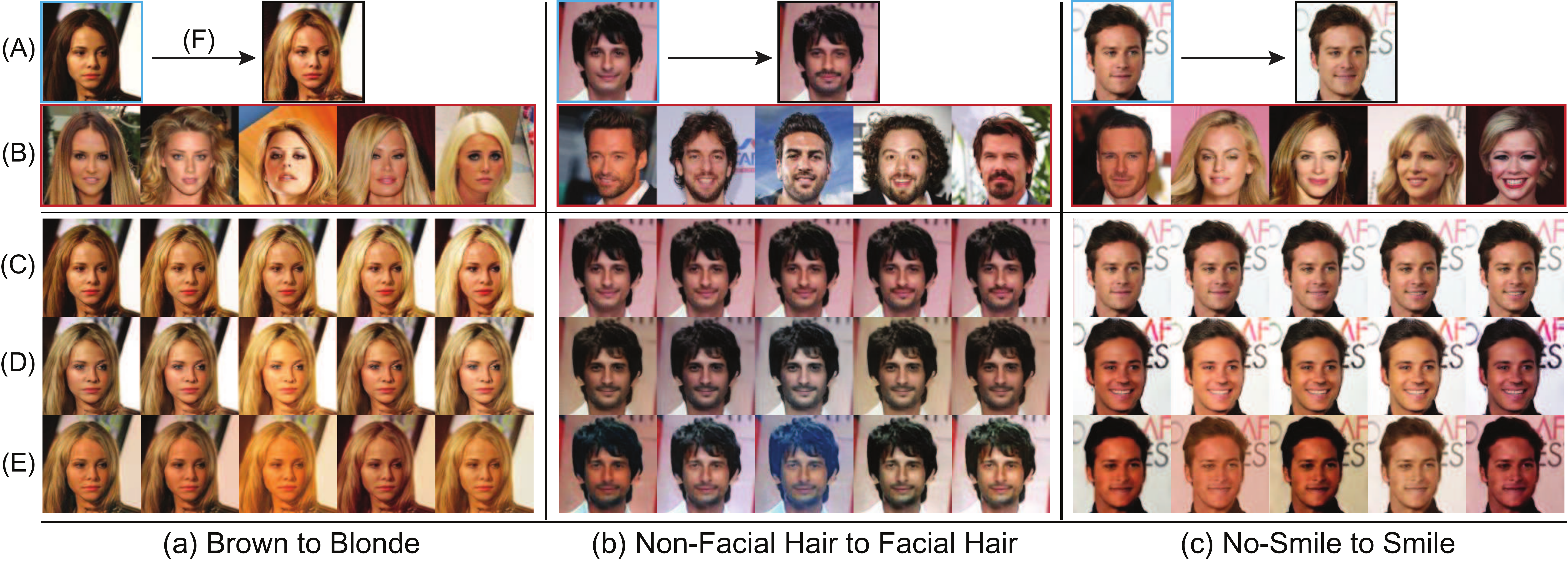}
  \caption{Comparison with the baseline models. Each row from the top represents~\textbf{(A)} an input image with~\textbf{(F)} an output of StarGAN~\textbf{(B)} given exemplars to the baseline models,~\textbf{(C)} LOMIT,~\textbf{(D)} DRIT, and~\textbf{(E)} MUNIT. The results demonstrates that LOMIT achieves the better performance than other baselines in the view of reflecting the given exemplar. Meanwhile, StarGAN, surrounded by the black box in the first row, is only able to generate an unimodal output.}
  \label{fig:comparison with baseline methods}
\end{figure*}

\paragraph{Mask regularization losses.}

We impose several additional regularization losses on local mask generation to improve the overall image generation performance as well as the interpretability of the generated mask. 

The first regularization is to minimize the difference of the mask values of those pixels having similar content information. This helps the local mask consistently capture a semantically meaningful region as a whole, e.g., capturing the entire hair region even when the lighting conditions and the hair color vary significantly within the exemplar. 
To this end, we design this regularization to minimize, as
\begin{equation*}\label{cd}
\begin{matrix}
    \mathcal{R}_{1}=
    \mathop{\mathbb{E}}\left[\sum_{i,j}\left[|(m\cdot{\vec{1}^{\,T}})-(\vec{1}\cdot{m^T})|\odot (\hat{c}\cdot{\hat{c}^{\,T}})\right]_{ij}\right],
\end{matrix}
\end{equation*}
where ${i=\{1,...,W\}}$, ${j=\{1,...,H\}}$, and $\vec{1}$ is a vector whose elements are all ones. Note that each of ${\{\vec{1}, m\}}$ is in ${\mathbb{R}^{WH\times 1}}$, and ${\hat{c}}$ is in ${\mathbb{R}^{WH\times C}}$, where $\hat{c}=\frac{c}{\lVert c\rVert}$. The first term is the distance matrix of all the pairs of pixel-wise mask values in ${m}$, and the second term is the cosine similarity matrix of all the pairs of $C$-dimensional pixel-wise content vectors. 
Note that we backpropagate the gradients generated by this regularization term only through ${m}$ to train the attention networks, but not through $\hat{c}$, which prevents the regularization ${R_1}$ from affecting the encoder ${E}$. 
The second regularization is to minimize the local mask region~\cite{Chen_2018_ECCV,pumarola2018ganimation}, i.e., 
\begin{equation*}\label{min}
    \mathcal{R}_{2}=\mathop{{}\mathbb{E}}{\lVert m\rVert_1},
\end{equation*}
which plays a role in encouraging the model to focus only on a necessary region involved in image translation.

\paragraph{Other losses.} 
To enable our framework can be trained while generating the best result, we use the image-level reconstruction losses~\cite{Zhu_2017,Huang_2018_ECCV,Lee_2018_ECCV} and the domain adversarial loss~\cite{arjovsky2017wasserstein,gulrajani2017improved}. We also adopt an auxiliary classifier~\cite{odena2016conditional,StarGAN2018} to cover multi-attribute translation with a single shared model, with which a user can choose single or multiple attributes to translate by manipulating a mask for an exemplar.

\section{Experiments}\label{sec:experiments}
In this section, we elaborate the settings and the results of our experiments. 

\subsection{Comparison with Baselines}\label{sub:Comparison with baselines}

In this section, we qualitatively and quantitatively compare LOMIT with MUNIT~\cite{Huang_2018_ECCV}, and DRIT~\cite{Lee_2018_ECCV} on CelebA dataset~\cite{liu2015deep}. We also compare LOMIT with AGGAN~\cite{mejjati2018unsupervised} on wild image dataset~\cite{Zhu_2017}. Finally, we additionally report the result of LOMIT on EmotioNet dataset~\cite{emotionet}.

\paragraph{Evaluation on CelebA dataset.}

As shown in Fig.~\ref{fig:comparison with baseline methods}, we compare our model with the baseline models using CelebA dataset~\cite{liu2015deep}, where we train the baselines based on publicly available model implementations. Each macro column from the left indicates the translation from (a) brown to blonde, (b) non-facial hair to facial hair, and (c) non-smile to smile. LOMIT shows an outstanding performance compared to the baseline models in both reflecting the distinct style of an exemplar while keeping the irrelevant region, such as the background and the face in the hair color translation, intact. 

Concretely, we observe that the noise in the style information extracted from the background is undesirably affecting the generated images in the case of MUNIT and DRIT (notably in the third column of both (a) and (b), and the first and the fifth columns of (c)), while LOMIT does not suffer from such influence. Besides, we also find that MUNIT and DRIT apply the style information to the irrelevant region of the input images, distorting the color and the tone of both the face and the background. It evidences that the mask for the exemplar should be properly incorporated and that LOMIT is superior to the compared models with regard to the accurate style application. These findings justify the initial motivation and the needs of the local masks along with the proposed HAdaIN module of LOMIT.

We additionally compare StarGAN~\cite{StarGAN2018} to verify the benefits of LOMIT. It demonstrates that StarGAN is only able to generate a unimodal output depending on an multi-hot input vector indicating a target attribute. On the other hand, LOMIT generates diverse outputs reflecting each corresponding exemplar.

\begin{table}[h]
\begin{center}
\begin{tabular}{cccc|c}
\toprule
& H \& S & F.H \& G & M \& Y & Avg. \\
\midrule
MUNIT & 31.52 & 44.81 & 37.78 & 38.04 \\
DRIT & 26.94 & 29.57 & 33.68 & 30.06 \\
\textbf{LOMIT} &\textbf{21.82} & \textbf{19.31} & \textbf{26.01} & \textbf{22.38} \\
\bottomrule
\end{tabular}
\end{center}
\caption{Comparisons of the \textbf{FID} in the target domain. Each of H\&S, F.H\&G, and M\&Y indicates hair colors (`Brown\_Hair', `Blonde\_Hair', `Black\_Hair') \& `Smiling', facial hair (`Mustache', `No\_Beard', `Goatee') \& `Male', and `Heavy\_Makeup' \& `Young', following the configurations in CelebA.}
\label{table:FID}
\end{table}

We also compare LOMIT with the baseline models using FID~\cite{heusel2017gans}, one of the renowned metrics for measuring the performance of generative models. Table.~\ref{table:FID} lists the comparison results. In all the class subsets, LOMIT generates images that are more diverse and of better quality than the other methods, as indicated by the lower scores. We believe this is attributed to the capability of LOMIT in applying the extracted style to the adequate region of the input image while keeping irrelevant regions intact. On the other hand, DRIT and MUNIT apply the style to unnecessarily large area of an input image, ending up with the distributions of generated images being far from the real distribution, as quantified by the high FID scores.

\paragraph{Evaluation on wild image dataset.}

\begin{figure}[t]
  \includegraphics[width=\linewidth]{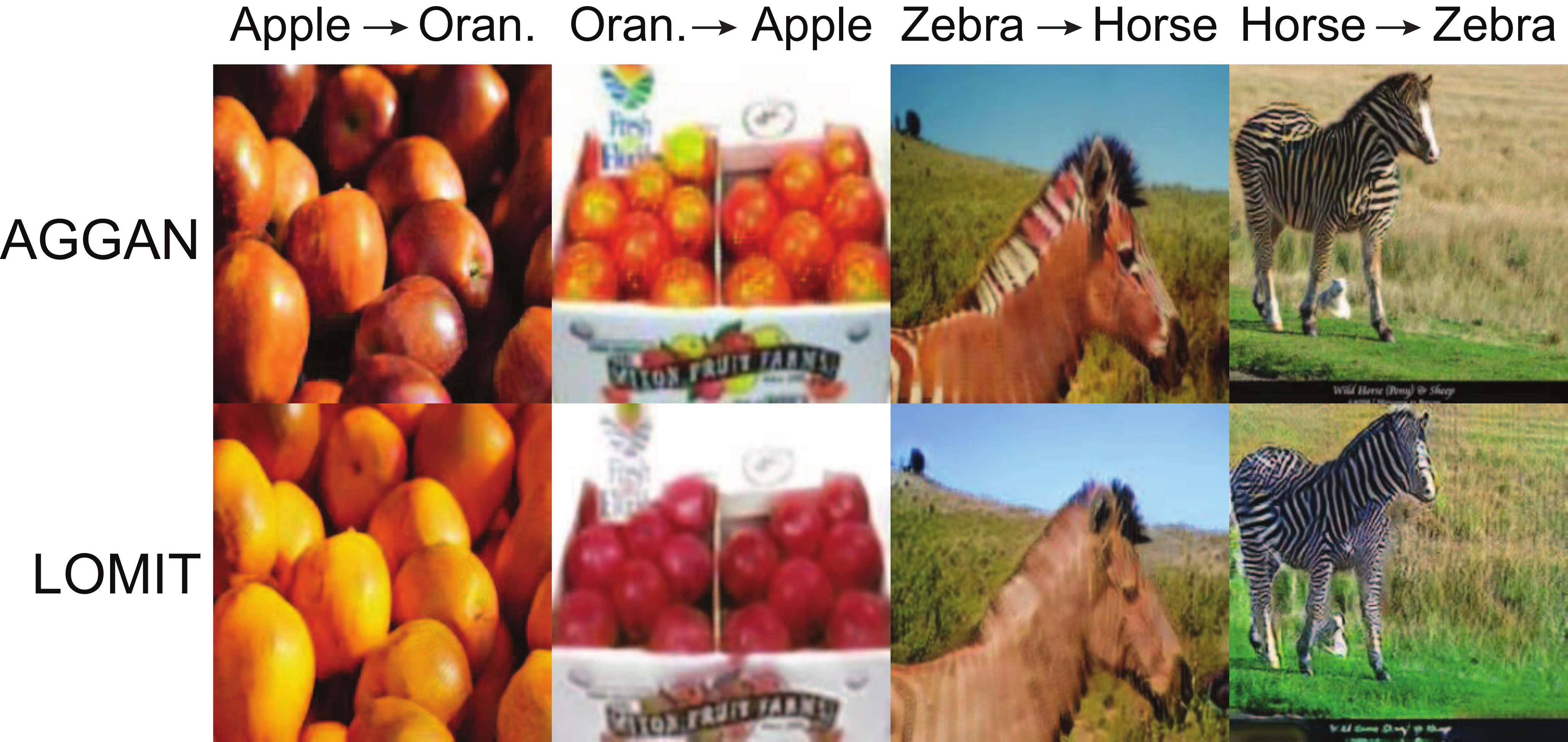}
  \caption{Comparison with AGGAN. Each row represents the results of AGGAN and LOMIT, and each column shows the different translation case. The used exemplar for each case is the input image of the opposite case.}
  \label{fig:comparison with AGGAN}
\end{figure}

In order to verify LOMIT can handle dataset having large variations within a class, we perform comparison experiments on wild image dataset~\cite{Zhu_2017} using the FID scores. For this experiment, we adopt AGGAN~\cite{mejjati2018unsupervised} as the state-of-the-art mask-based image translation method. We use the official source code with minor modifications on the architecture to take an exemplar as another input. 

\begin{table}[h]
\begin{center}
\begin{tabular}{ccccc|c}
\toprule
& A$\rightarrow$O & O$\rightarrow$A & H$\rightarrow$Z & Z$\rightarrow$H & Avg.\\
\midrule
AGGAN & 170.83 & 110.61 & 105.27 & 97.25 & 120.99 \\
\textbf{LOMIT} & \textbf{124.31} & \textbf{91.21} & \textbf{49.60} & \textbf{89.74} & \textbf{88.72} \\

\bottomrule
\end{tabular}
\end{center}
\caption{FID score comparison with AGGAN. Each of A,O,H, and Z denotes Apple, Orange, Horse, and Zebra, respectively.}
\label{table:comparison with AGGAN}
\end{table}

For the four tasks of horse2zebra, zebra2horse, apple2orange, and orange2apple, the overall FID scores of LOMIT outperform AGGAN as shown in Table.~\ref{table:comparison with AGGAN}. We also verify the superior performance of LOMIT via qualitative comparisons with AGGAN as illustrated in Fig.~\ref{fig:comparison with AGGAN}. For example, the results of AGGAN in the first and the second columns show the incorrect translation results because the orange and the red are insufficiently colored. We believe that those results basically come from the use of the exemplar-mask. AGGAN cannot specify a region in an exemplar to be used as the style, while LOMIT is capable of refining the style by selecting essential regions of the exemplar, thus LOMIT can produce better results. 


Moreover, LOMIT can perform well even when a given exemplar contains multiple instances. For example, for the translation from zebra to horse, if a given exemplar is composed of multiple horses consisting both white horses and brown horses, the baseline method cannot specify which horse to be applied as a style in a translation. On the other hand, LOMIT tackles the problem and enables to choose which instance to be referenced as a style via masking on the exemplar.

\begin{figure}[t]
  \includegraphics[width=\linewidth]{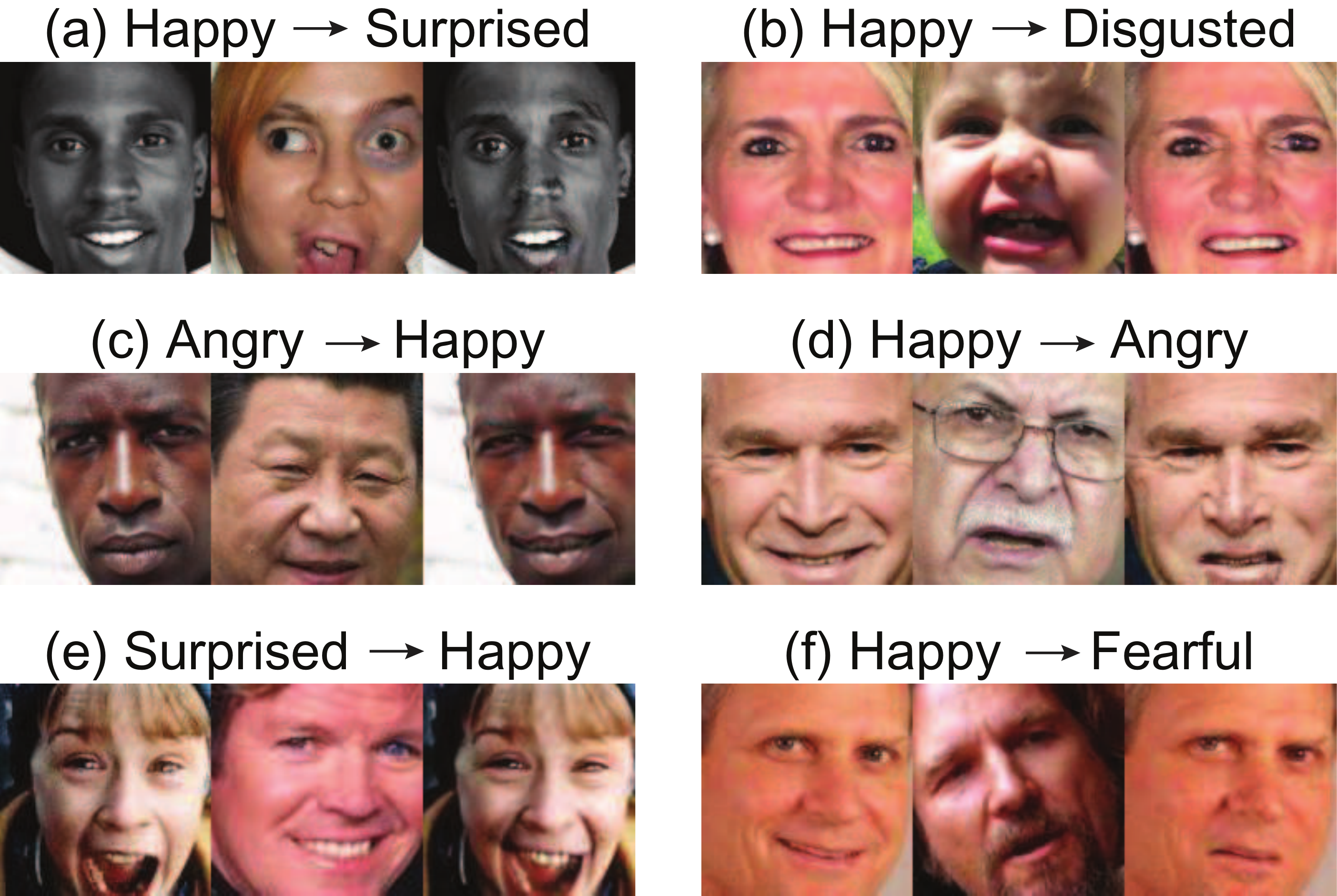}
  \caption{The result of action unit translation using EmotioNet dataset. The figure shows LOMIT can transfer various AUs from a given exemplar, such that it results in a change of an emotion following the given exemplar.}
  \label{fig:emotion}
\end{figure}

\paragraph{Experiment on EmotioNet dataset.} Fig.~\ref{fig:emotion} shows the results for the action unit (AU) translation, using EmotioNet dataset~\cite{emotionet}. For the training, we use all available AUs (1, 2, 4, 5, 6, 9, 12, 17, 20, 25, 26, and 43) as a training label (for the multi-attribute translation loss), so that the model can be trained for translating multi-AUs from the exemplar. Each triplet is composed of an input image, an exemplar, and a translated output. For example, the input of (a), containing AUs 12, 25 (Happy) takes the exemplar whose AUs are 1, 2, 25, 26 (Surprised). The translated output demonstrates that it preserves the identity of the input image while properly transferring the AUs of the exemplar. From the results, we verify that although a number of AUs sparsely distributed all over the face, LOMIT can perform the elaborate translation based on the local masks.


\begin{figure*}
\centering
\begin{minipage}[b]{0.48\textwidth}
\includegraphics[width=\linewidth]{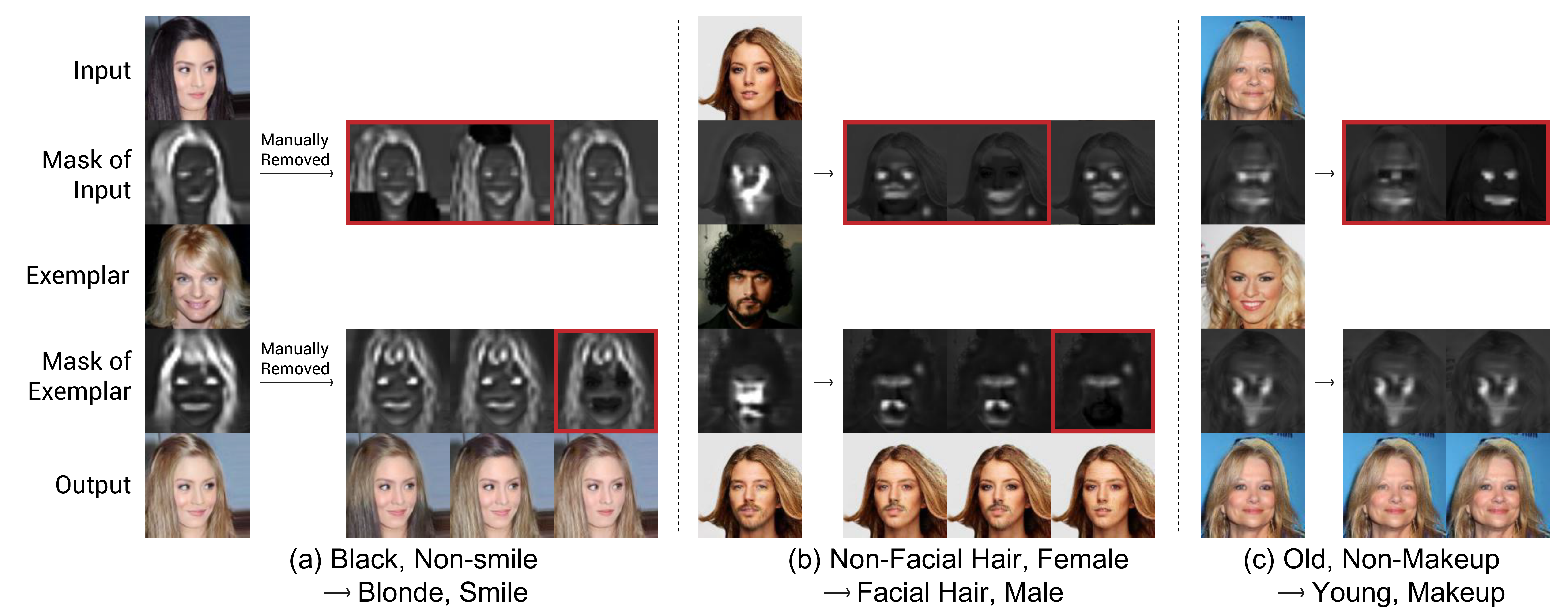}
\caption{A user can specify the regions from either or both an input image and an exemplar by manipulating masks.}\label{fig:interaction figure}
\end{minipage}\qquad
\begin{minipage}[b]{0.48\textwidth}
\includegraphics[width=\linewidth]{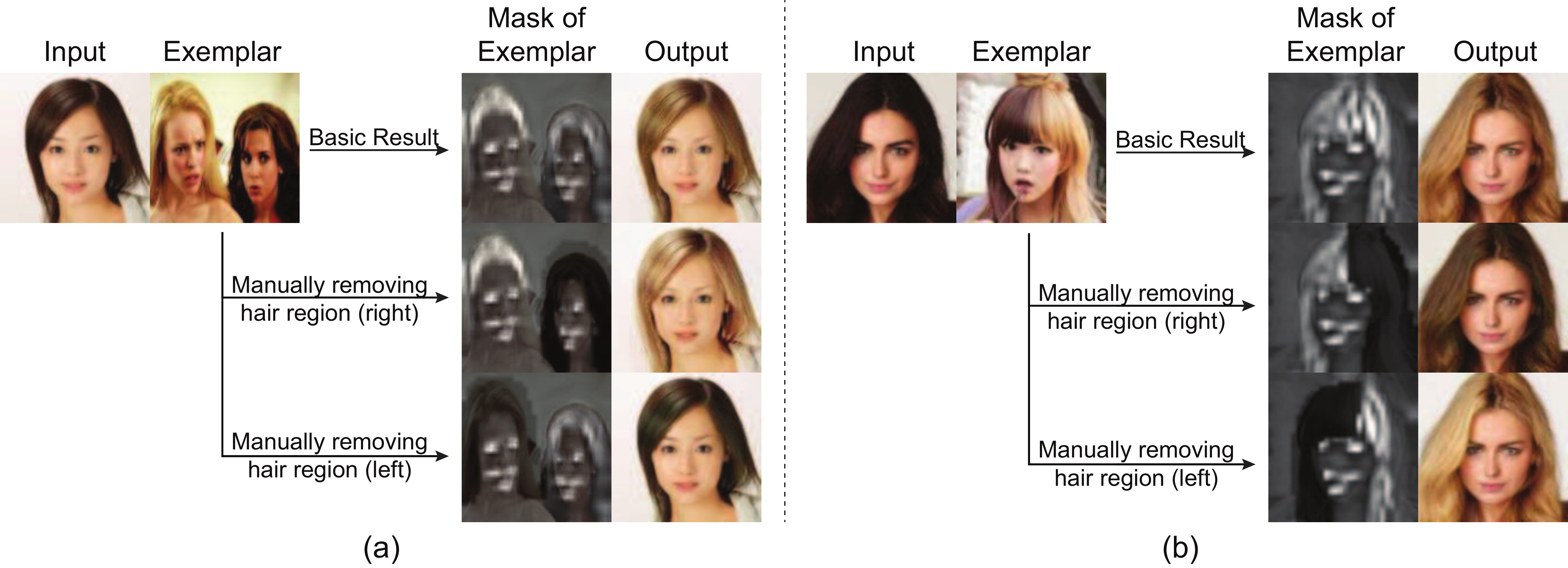}
\caption{Practical examples that require manipulations of users on the exemplar mask in the midst of the translation process.}\label{fig:style mask interaction}
\end{minipage}
\vspace{-0.5cm}
\end{figure*}





\subsection{User Intervention using LOMIT}\label{sub:examples on practical use.}

\paragraph{Various translation examples via user interaction.}
Fig.~\ref{fig:interaction figure} demonstrates a multifarious applicability of LOMIT via human interaction. It is shown that a user can specify where to translate as well as what to be transferred by manipulating the masks of an input and an exemplar. Masks with red outlines are the ones manually removed. The leftmost column of each macro column shows the result without any modification, and the other columns of each macro column indicate translation results after modifying the mask of the input or the exemplar. Users can modify the input mask to adjust where to translate, while one could change the exemplar mask to decide what to translate. For example, the second and the third columns in Fig.~\ref{fig:interaction figure}(a) are the results from a modification of the input-mask. From those results, it is verified that a user can choose where to translate through opting the region to apply the extracted style. Meanwhile, the fourth column shows the result from a modification of the exemplar-mask. It maintains the non-smile attribute of the input image by removing the regions of the eyes and the mouth of the exemplar-mask, which contains the smile attribute information. This demonstrates that a user can choose which style to transfer during a translation procedure, so that the learned attributes to be translated during training can be selectively transferred via user-intervention on the exemplar-mask.

\paragraph{Necessity of modifying exemplar-mask.}
We believe this approach bears great potentials in diverse computer vision applications by allowing the user to fine-control the target region for the translation as well as the style of the exemplar. In particular, the technique can be effectively applied when the different styles in the same attribute co-exist in an exemplar. Practically, as illustrated in Fig.~\ref{fig:style mask interaction}(a) and (b), there can be numerous cases that an exemplar contains distinct styles within a single attribute. For example, the exemplar of Fig.~\ref{fig:style mask interaction}(a) contains two women having different hair colors. It indicates that the target style extracted from the exemplar can be vague. The basic result represented in the first row shows the mixed color of brown and blonde, while the results from the second and the third rows respectively represent blonde and brown hair color because their exemplar-masks are modified in order to specify the hair color. By explicitly specifying the exact style to transfer, a user can designate a concrete target style, and the model can conduct an appropriate translation reflecting what the user wants.


\begin{figure}[b]
  \includegraphics[width=\linewidth]{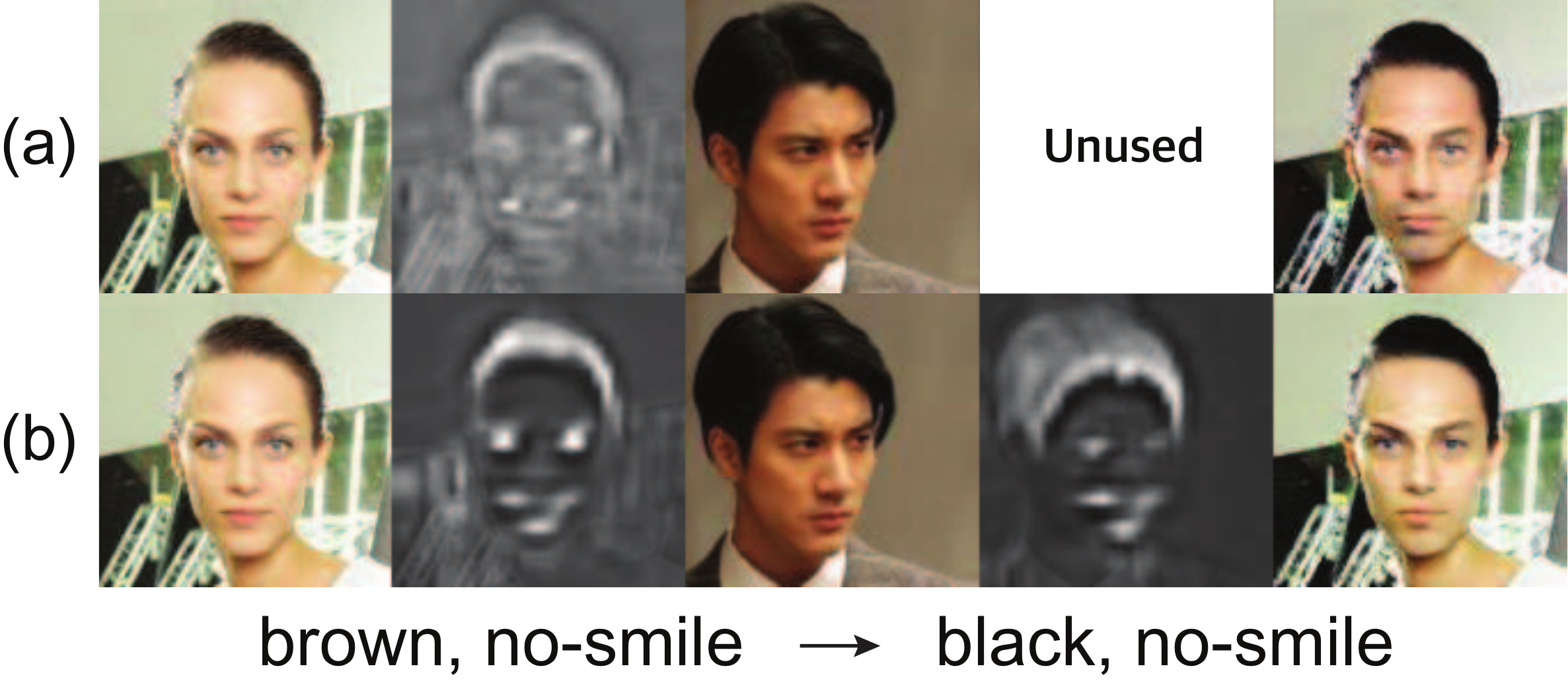}
  \caption{Effects of the style mask. Each row from the top represents (a) $\text{LOMIT}^{--}$ and (b) LOMIT.}
  \label{fig:stylemask analysis}
\end{figure}

\subsection{Analysis on the exemplar mask}


We demonstrate the effects of the mask for the exemplar in Fig.~\ref{fig:stylemask analysis}. From the left, each column of the figure is composed of the input image, the corresponding input mask, the exemplar, its corresponding mask, and the output. The first row shows the result trained without incorporating the mask for the exemplar ($\text{LOMIT}^{--}$), meaning that the style encoder encodes the entire regions of the exemplar as the foreground style, as well as that of the input as the background style. As illustrated in the second and fourth columns in the figure, the style masks of LOMIT (row (b)) specify regions more clearly than $\text{LOMIT}^{--}$ (row (a)). It reveals that the style mask is effectively regulating the model to distinguish the region of interest while minimizing the distortion of the irrelevant regions. Concretely, as can be seen in the image in the fifth column of (a), the area surrounding a mouth and the sculpture in the left is affected by the style, due to the mask excessively specifying the regions. Besides, the reddish face of the image demonstrates the extracted style from the exemplar includes extraneous regions on the translation, because a skin tone of a face is irrelevant information in the translation of (brown, no-smile ${\rightarrow}$ black, no-smile). On the other hand, the image in the fifth column of (b) does not only maintain the irrelevant region, but also reflects the style to the relevant regions. That is, by exploiting the mask for the exemplar, we can achieve a better translation result.

\section{Conclusion}
In this work, we addressed the problem of what and where to translate for unpaired image-to-image translation. We proposed a local mask-based translation model called LOMIT, where the attention networks generate the mask of an input image and that of an exemplar. The mask of the exemplar determines what style to transfer by excluding irrelevant regions and extracting the style from only relevant regions. The other mask of the input determines where to transfer the extracted style. That is, it captures the regions to apply the style while maintaining an original style in the rest (through our highway adaptive instance normalization). As future work, we plan to extend our model to other diverse domains of data, such as ImageNet~\cite{imagenet_cvpr09} and MSCOCO~\cite{lin2014microsoft}. We will also extend our approach to video translation to improve the consistency of the translated results of consecutive frames.




\clearpage

\bibliographystyle{named}
\bibliography{ijcai20}

\end{document}